# Patch-based Fake Fingerprint Detection Using a Fully Convolutional Neural Network with a Small Number of Parameters and an Optimal Threshold

Eunsoo Park, Xuenan Cui, Weonjin Kim, Jinsong Liu, and Hakil Kim

**Abstract**—Fingerprint authentication is widely used in biometrics due to its simple process, but it is vulnerable to fake fingerprints. This study proposes a patch-based fake fingerprint–detection method using a fully convolutional neural network with a small number of parameters and an optimal threshold to solve the above-mentioned problem. Unlike the existing methods that classify a fingerprint as live or fake, the proposed method classifies fingerprints as fake, live, or background, so preprocessing methods such as segmentation are not needed. The proposed convolutional neural network (CNN) structure applies the Fire module of SqueezeNet, and the fewer parameters used require only 2.0 MB of memory. The network that has completed training is applied to the training data in a fully convolutional way, and the optimal threshold to distinguish fake fingerprints is determined, which is used in the final test. As a result of this study's experiment, the proposed method showed an average classification error of 1.35%, demonstrating a fake fingerprint–detection method using a high-performance CNN with a small number of parameters.

## I. INTRODUCTION

Recently, discussion on fake fingerprint detection has been revitalized owing to the increased use of fingerprint authentication systems in mobile devices and incidents such as record-access manipulation using fake fingerprints. Because fingerprints can easily be faked using silicon, gelatin, clay, etc., fake fingerprint–detection systems need the ability to detect fake fingerprints in case safety and security are paramount. Fig. 1 compares a live fingerprint and a fake fingerprint for the same finger by using the Biometrika sensor in LivDet2011 [1]. Thus, integrated feature design that can conclusively tell fake fingerprints is very difficult, because the features appear a little bit different, depending on the material used to produce a fake fingerprint, as shown in Fig. 1. Existing fake fingerprint–detection methods can be classified into software-based and hardware-based methods. Hardware-based methods have the disadvantage of requiring an additional sensor to verify liveness.

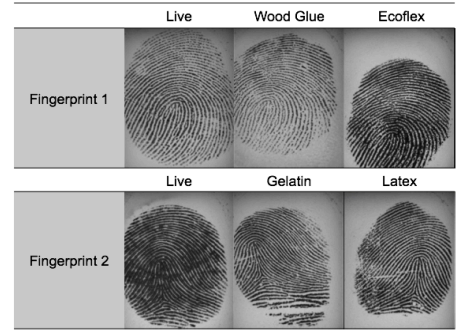

Fig. 1. Visual comparison between live and fake fingerprints of the same finger.

On the other hand, software-based methods do not need additional sensors, but have the disadvantage of lower performance in distinguishing fake fingerprints, compared to hardware-based methods [2].

Convolutional neural networks (CNNs), which have been used widely in image processing in recent years, show the best performance in the classification and detection of images, and are applied in many industrial fields. If the high classification performance of the CNN method can be applied to the detection of fake fingerprints, software-based fake fingerprint–detection methods will have room to improve their relatively low detection performance.

However, a CNN generally requires a lot of memory and processing time because they have many parameters. This factor makes it difficult to apply fake fingerprint–detection methods using a CNN in real life. Furthermore, the size of fingerprint images must be changed in accordance with the fixed input size of the CNN, which causes loss and distortion in the fingerprint image. For the universal application of fake fingerprint detection using a CNN with high classification performance, it must be possible to process the input fingerprint images without changing their size and to have a small number of parameters.

This study proposes a patch-based fake fingerprint–detection method using a fully convolutional neural network with a small number of parameters and an optimal threshold to solve the

This work was supported by an Institute for Information & communications Technology Promotion (IITP) grant funded by the Korean government (MSIP) (No.R0190-15-2007, Development of film type fingerprint sensor module and privacy protective application software for smart devices).

E. Park, X. Cui, W. Kim, J. Liu, and H. Kim are with the Information and Communication Engineering Department, Inha University, 100 Inharo, Incheon, South Korea (e-mail: espark@inha.edu, xncui@inha.ac.kr, wjkim@inha.edu, liujs@inha.edu, hikim@inha.ac.kr).

above-mentioned problems. The proposed method applies the following constraints to the design of such a CNN.

- It does not convert the size of the input fingerprint images in order to avoid information loss.
- It is designed to receive various sizes of fingerprint images as input.
- The number of parameters for the CNN is small.

In order to accept various sizes of fingerprint images without converting them, a method of detecting fake fingerprints using patches instead of the full fingerprint image is proposed. If patches are used instead of the full fingerprint when detecting fake fingerprints, the following advantages can be obtained.

First, it is possible to construct networks independent of fingerprint size by constructing them according to the patch size set by the designer instead of the fingerprint image size. Second, the amount of training data can be increased, and features unrelated to position can be learned because they are learned in patch units. Third, it is easy to determine local characteristics of fake fingerprints through visualization of the results processed in a patch unit. Fourth, networks with less information loss can be constructed by reducing the pooling count of the CNN because patches are smaller than fingerprint images. Fifth and last, this method is applicable to devices such as mobile sensors that acquire only a specific area of a fingerprint.

Studies have been conducted on fake fingerprint detection applying a CNN while using patches of the fingerprint [3], [4], [5]. The proposed method has the following differences from those studies.

- Fingerprints are classified into live, fake, and background, instead of simply into the two classes of live and fake, and therefore, preprocessing (such as segmentation) is integrated into the CNN.
- Unlike the existing methods that use only one patch size, 32×32, 48×48, and 64×64 patch sizes were experimented with, and the appropriate patch size for the proposed network was determined.
- The fully convolutional neural network method can produce better results than an existing method that determines a fake fingerprint by dividing the fingerprint into non-overlap patches and applying the CNN to each patch.
- Determining a fake using the optimal threshold score obtained from the training data provides better performance than an existing method that determines a fake through voting after the fake of each patch is determined.
- A CNN model that maintains high performance with few parameters has been proposed through a design that restricts the number of parameters.
- Local characteristics of fake fingerprints can be visualized using the results of fully convolutional neural networks.

Based on the experiment results, the proposed method showed an average detection error of 1.35% for LivDet fake fingerprint data [1], [6], [7] with 536,143 parameters (approximately 2.0 MB), demonstrating the possibility of applying high-performance, fully convolutional neural networks with a small number of parameters to fake fingerprint detection. Furthermore, the proposed method can process any size of fake fingerprint because it uses patches, which are not affected by input fingerprint size, and there is no fully connected layer inside the network. Another advantage is that no preprocessing, such as segmentation, is required because the processed results are classified into fake, live, and background.

This paper is organized as follows. Section 2 introduces studies related to fake fingerprint detection. Section 3 explains the proposed method. Section 4 describes the experiment method and results. Finally, Section 5 outlines the conclusions and discusses future research directions.

## II. RELATED RESEARCH

Research on fake fingerprint detection started in 1998 by D. Willis and M. Lee who experimented with six sensors to find out how robust each sensor was against fake fingerprint attacks, revealing that four of them were vulnerable [8]. According to Coli et al. [9], fake fingerprint–detection methods can be divided into hardware-based and software-based, as follows.

- Hardware-based methods use additional hardware to extract physical features of the human body. They show more accurate detection performance, compared to software-based methods, but the disadvantage is high cost because sensors are added. Typical hardware-based methods include using blood pressure in the fingers [10], using the transformation of skin [11], and using skin odor [12].
- Software-based methods detect fake fingerprints using software algorithms. They are cheaper than hardware-based methods because no additional hardware is necessary. Most algorithms use physical data, like size, density, and continuity of fingerprint ridges. Recently, the application of a CNN has also been researched.

The method proposed in this study is a software-based fake fingerprint–detection method. For related research, therefore, studies on typical software-based fake fingerprint–detection methods and methods applying a patch-based CNN (which are similar to the proposed method) are examined.

Nikam and Agarwal [13] proposed a method that combines a local binary pattern (LBP) and a wavelet transform. The LBP histogram is applied to texture analysis, and the wavelet transform is applied to analysis of the frequency characteristics and directions of ridges. Nikam and Agarwal carried out further research and proposed an approach to extracting texture features by applying a wavelet and a gray-level co-occurrence matrix (GLCM) [14]. These approaches used principal component analysis and sequential forward feature selection to reduce the dimensions of feature sets.

Moon et al. [15] observed that the surface of fake fingerprints appears rougher than live fingerprints in a high-resolution camera. They used sensors that have a resolution of 1000 DPI, whereas general fingerprint sensors are 500 DPI. They divided the image into areas of fixed size because the image is too large. Their method was to use the remaining noise data after applying a noise removal algorithm. The standard deviation of this information is used to distinguish between fake and live fingerprints.

Coli et al. [16] used the fact that the fine features of live fingerprints do not appear clearly on fake fingerprints due to the rough surface and discontinuous ridges, and they proposed a method of classifying fake fingerprints after applying a Fourier transform to fingerprint images. To measure specific frequencies only, they defined high-frequency energy and used it for analysis.

Marasco and Sansone [17] proposed a fake fingerprint–detection method using texture features. Texture features include features that are generated through signal processing, such as the size of sweat glands and the noise of the fingerprint, statistical features such as variance and amount of information, and grey-level features of images. Galbally et al. [18] used a similar approach. They extracted texture features using a Gabor filter and used them for fake fingerprint detection. Gottschlich et al. [19] proposed and applied histograms of invariant gradients (HIG) by improving histogram of oriented gradients (HOG) and scale invariant feature transform (SIFT), transforming them in line with the textures of fingerprint ridges.

Ghiani et al. [20] applied local phase quantization (LPQ), which is robust against rotation, by improving the LBP. LPQ is mainly used in the analysis of low-frequency components, based on the observation that a low-frequency component analysis contains good features for distinguishing between fake and live fingerprints. Gragnaniello et al. [21] demonstrated this better performance by applying the Weber Local Descriptor (WLD) and LPQ together.

Jia et al. [22] devised a method of determining the liveness of fingerprints, which they called Multi-Scale Block Local Ternary Patterns, which uses features calculated through comparison between the average values of a specific area and the pixel values of the fingerprint area. Furthermore, Jia et al. extracted features by applying to fingerprint images the size change of a filter used in LBP and various linear filters, and then applying LBP again to the results. They applied these features to fake fingerprint images and proved the possibility of fake fingerprint detection at a higher performance level, compared to other studies [23].

Among the fingerprint detection methods using CNNs, Nogueira and colleagues [24], [25] and Marasco et al. [26] used 224×224 or 227×227 images (used in existing image classification) for network inputs instead of patches. Nogueira and colleagues [24], [25] used the transfer learning method to apply the VGG-19 layer model [27] to fake fingerprint detection. To use fingerprint images as input for the VGG model, they are reduced according to their ratio and cut to a fixed input size, which can cause information loss from the images. Marasco et al. [26] experimented with fake fingerprint–detection performance by using transfer learning with the parameters of GoogLeNet [28] in a manner similar to Nogueira et al [25]. Furthermore, they applied a one-shot learning method that compares the distance between fake and live fingerprints using a Siamese network, but its detection performance was not higher than the GoogLeNet method. The parameter counts of the VGG-19 model and GoogLeNet, which were used by Nogueira et al. and Marasco et al., were around 140 million and 6 million, respectively, which are greater than the proposed method, which is 0.54 million.

Studies that applied a CNN to patches of fingerprints include Park et al. [5], Wang et al. [3], and Jang et al. [4]. Park et al. randomly extracted 11 patches at 96×96 around the segmented fingerprint areas, classified the patches using a CNN with three convolution layers and one fully connected layer, and then finally determined whether the fingerprint is fake or not by voting. However, the number of data used in the experiment was too small, because only the results of the Identix sensor in LivDet 2009 [29] were presented, and the number of network parameters is around 16 million, which is large. Wang et al. applied segmentation to the input fingerprint images and extracted patches at 32×32 from the images. Then, each patch was evaluated using a CNN consisting of four convolution layers and one fully connected layer, with the final result determined through voting. The number of parameters in Wang et al.'s CNN structure is around 0.5 million, which is very small. Unlike Wang et al.'s method, that of Jang et al. applies histogram equalization to the fingerprint images and extracts 16×16 patches. This method is similar to Wang et al.'s method in that four convolutions and two fully connected layers are used, and a final voting method is applied. However, the number of parameters is around 8.6 million, which is rather large, because two fully connected layers with 2048 neurons are used. Wang et al.'s method showed a 0% detection error in three of four sensors of the LivDet2011 data [1] and a 0.9% error rate, on average—the highest performance among the methods that have been proposed until now. Jang et al.'s method also showed very high performance at a 0.2% misdetection rate on ATVS data [30]. However, in Wang et al.'s and Jang et al.'s papers, no information about the composition of validation data exists, and only the test performance was presented. Wang et al.'s and Park et al.'s methods have a disadvantage, because segmentation must be applied to extract patches, and Jang et al.'s method has another disadvantage because it uses histogram equalization as a preprocessing method. Furthermore, it can be verified through the proposed method that extracting results from all areas by applying a fully convolutional network produces better performance than estimating the detection results of non-overlap patches, which were used for testing by Wang et al.'s and Jang et al.'s methods. It was also verified through this study that deriving the final result using the proposed optimal threshold produces better performance than the voting method when making the final judgment about a fingerprint.

## III. PROPOSED METHOD

The overall structure of the proposed method is shown in Fig. 2. First, segmentation is applied to train the CNN, and patches of live, fake, and background images are extracted based on the area. The extracted patches are used in training the CNN, and the learned CNN is a fully convolutional structure used to find the optimal threshold. After the images classified as background are removed from the results of the fully convolutional network (FCN), the optimal threshold for distinguishing between fake and live fingerprints is found from the training data. In the test process, the trained FCN is applied, and it determines whether the fingerprint is fake or not by

$$M = \frac{1}{width \times height} \sum_{i=0}^{width} \sum_{j=0}^{height} AVG_{i,j} \quad (3)$$

$$V = \frac{1}{width \times height} \sum_{i=0}^{width} \sum_{j=0}^{height} VAR_{i,j} \quad (4)$$

$$AVG_{i,j} < M \text{ and } VAR_{i,j} > V \quad (5)$$

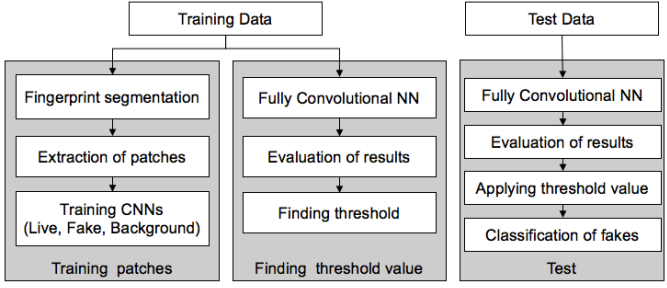

Fig 2. The structure of the proposed fully convolutional neural networks with optimal threshold value for patch based fingerprint liveness detection.

$I$ in (1) and (2) denotes a fingerprint image; $AVG_{i,j}$ denotes the average of a block with a size of $bSize \times bSize$ in the image coordinate, $I(i, j)$. As mentioned above, $bSize$ was determined to be 9. Likewise, Equation (2) denotes the variance of a block at $bSize \times bSize$ in image coordinate $I(i, j)$. The average and variance of blocks corresponding to the entire image can be expressed with equations (3) and (4). The fingerprint images $I(i, j)$ corresponding to the block satisfying (5) using these reference values are classified as a fingerprint area; otherwise, they are classified as background. Equation (5) was created under the assumption that the fingerprint area generally has a low average value and a high variance.

The training patches are extracted from the segmentation results. First, fingerprint images are divided into grid-shaped patches that do not allow overlapping. If the segmented area in each divided patch is lower than the specified ratio, the patch is classified as background. For example, a 32×32 patch has a total of 1024 pixels, and if 800 pixels are segmented as a fingerprint, the ratio of the segmented area becomes 800/1024=0.78, but it is classified as background if this ratio is smaller than 0.4. If the value for judging the ratio of the segmented area is $R$, the $R$ value for the 32×32 patch is set to 0.4. The $R$ value changes, depending on the patch size, to $R \cdot (1/2)$ for 48×48 patches and to $R \cdot (1/3)$ for 64×64 patches. The $R$ value changes adaptively, because more fingerprint areas are included in the background class as the patches become larger.

In general, when a fingerprint is divided into patches, the number of background classes is very small. Therefore, the number of background patches is matched to the number of other patches through data augmentation. Table I lists the number of acquired patches by size. The first number in the comparing the result with the optimal threshold. Each process is described in detail in the following sections.

### A. Acquisition of training patches

The patches acquired from the training data must have a background class, but a general fake fingerprint database does not have a background class. Therefore, we must directly create a background class from the acquired patches. It is possible to visually check each patch to assign the background class, but in the proposed method, automatic classification of the background by image segmentation is used. Fingerprint segmentation is carried out in the following sequence. First, the fingerprint image is divided into blocks sized 9×9. After the average and variance values of each block area are calculated from the entire fingerprint image, all the pixels of the fingerprint image are classified into background and fingerprint areas based on these two values. This process can be expressed as follows:

$$AVG_{i,j} = \frac{1}{bSize^2} \sum_{k=0}^{bSize-1} \sum_{l=0}^{bSize-1} I(i + k, j + l) \quad (1)$$

$$VAR_{i,j} = \frac{1}{bSize^2} \sum_{k=0}^{bSize-1} \sum_{l=0}^{bSize-1} (I(i + k, j + l) - AVG_{i,j})^2 \quad (2)$$

TABLE I
Number of patches acquired by data type.

| Patch Size | | 32×32 | | | 48×48 | | | 64×64 | | |
|---|---|---|---|---|---|---|---|---|---|---|
| LivDet Sensor | | Live | BG(Ori/Aug) | Fake | Live | BG(Ori/Aug) | Fake | Live | BG(Ori/Aug) | Fake |
| 11 | Bio | 59271 | 250 / 58317 | 58843 | 28403 | 87 / 27315 | 27670 | 15205 | 36 / 13957 | 14250 |
| | Dig | 69986 | 392 / 63753 | 63964 | 35234 | 128 / 32002 | 32375 | 22041 | 46 / 19940 | 20338 |
| | Ital | 71645 | 1515 / 72571 | 75715 | 36274 | 611 / 36545 | 38249 | 22301 | 311 / 22313 | 23317 |
| | Sag | 77561 | 325 / 75780 | 76580 | 37847 | 93 / 36900 | 37572 | 22542 | 30 / 21468 | 22241 |
| 13 | Bio | 56819 | 186 / 56027 | 58112 | 27395 | 58 / 26844 | 27558 | 14744 | 17 / 13983 | 14597 |
| | Ital | 67588 | 1588 / 64994 | 64394 | 34291 | 644 / 32764 | 32839 | 21237 | 329 / 20331 | 20367 |
| 15 | Bio | 273971 | 4655 / 199780 | 196779 | 129129 | 1852 / 97956 | 96900 | 76787 | 1017 / 58846 | 58766 |
| | Cro | 124928 | 2496 / 124928 | 122230 | 62591 | 1014 / 61667 | 61461 | 38662 | 550 / 38373 | 38041 |
| | Dig | 47775 | 100 / 46724 | 50171 | 22928 | 26 / 23777 | 23777 | 12231 | 8 / 11274 | 12758 |
| | Gre | 72500 | 952 / 61679 | 61604 | 36511 | 400 / 31494 | 31538 | 21620 | 167 / 19430 | 19441 |

background (BG) column in Table I indicates the number of acquired background patches, and the second number indicates the total number of patches acquired through data augmentation. The greater the size of the patches, the more insufficient the number of acquired background patches becomes. To increase the number of background classes, random rotation, scale, and translation changes are used in data augmentation.

### B. Patch-based CNN model

The proposed CNN model uses the Fire module of SqueezeNet [31]. The Fire module consists of the Squeeze layer, which reduces the number of input channels using a small number of 1×1 convolution layers, and the Expand layer, which increases the number of channels of results again using 1×1 and 3×3 convolutions, with the results of the Squeeze layer as input. This method is called a bottle-neck structure. The Expand layer of the Fire module also has 1×1 convolutions in an effort to further reduce the number of parameters.

The structure of the proposed network using the Fire module is shown in Fig. 3. To represent the actual shape of the filters, the filters of the Squeeze layer in the Fire module are expressed as long and small numbers, and the filters of the Expand layer are expressed as short and many. The number after the Fire module in the network architecture in Fig. 3 indicates the number of filters of the Expand layer. The number of filters of the Squeeze layer in the Fire module is 0.125 times that of the Expand layer, and this ratio is called Squeeze ratio. The ratio of the number of 1×1 and 3×3 filters was set to 0.5 in the Expand layer. This Squeeze ratio and the ratio of the number of filters of the Expand layer were used for every Fire module. In other words, the Expand layer of fire5-256 in Fig. 3 has a total of 256 filters which consists of 128 1×1 and 3×3 filters. The number of 1×1 filters of the Squeeze layer is 32, or 0.125 times the number of filters of the Expand layer, which is 256.

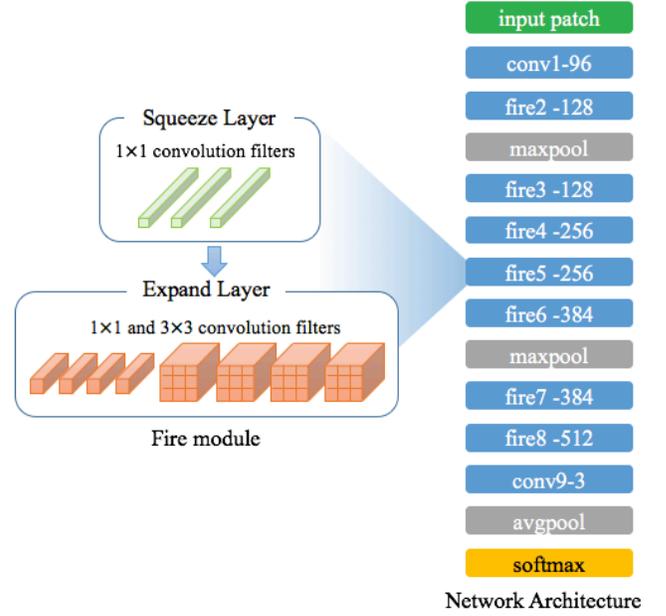

Fig 3. Proposed CNN structure using the Fire module.

The proposed CNN structure applies pooling only three times. For the activation function in the proposed network, Leaky ReLU [32], as shown in (6), was applied, and the $a$ value was set to 0.3.

$$f(x) = \max(x, ax) \qquad (6)$$

Furthermore, the batch normalization layer [33] was added before applying the activation function in every layer. During training, Dropout [34] was applied at a probability of 0.5 after the fire8-512 layer in Fig. 3. Table II shows the structure of the proposed CNN module in more detail. The filter structure is identical, regardless of the size of input patches, except that the filter sizes of avgpooling (which is the last one in Table II) are 7×7, 11×11, and 15×15, in line with the input tensor size. The number of parameters in Table II was calculated without batch normalization parameters, and the number is separately indicated when batch normalization parameters are included. The total number of parameters of the proposed network with this structure is 536,143, which is approximately 2.0 MB if one parameter is four bytes.

TABLE II
PATCH-BASED CNN NETWORK STRUCTURE

| Layer name / type | Output size | | | Filter size / stride (if not a fire layer) | depth | $s_{1×1}$ (#1×1 squeeze) | $e_{1×1}$ (#1×1 expand) | $e_{3×3}$ (#3×3 expand) | # of parameters |
|---|---|---|---|---|---|---|---|---|---|
| input image | 32×32×1 | 48×48×1 | 64×64×1 | | | | | | |
| conv1 | 32×32×96 | 48×48×96 | 64×64×96 | 3×3/1 (×96) | 1 | | | | 960 |
| fire2 | 32×32×128 | 48×48×128 | 64×64×128 | | 2 | 16 | 64 | 64 | 11,920 |
| maxpool2 | 15×15×128 | 23×23×128 | 64×64×128 | 3×3/2 | 0 | | | | |
| fire3 | 15×15×128 | 23×23×128 | 31×31×128 | | 2 | 16 | 64 | 64 | 12,432 |
| fire4 | 15×15×256 | 23×23×256 | 31×31×256 | | 2 | 32 | 128 | 128 | 45,344 |
| fire5 | 15×15×256 | 23×23×256 | 31×31×256 | | 2 | 32 | 128 | 128 | 49,440 |
| fire6 | 15×15×384 | 23×23×384 | 31×31×384 | | 2 | 48 | 192 | 192 | 104,880 |
| maxpool6 | 7×7×384 | 11×11×384 | 15×15×384 | 3×3/2 | 0 | | | | |
| fire7 | 7×7×384 | 11×11×384 | 15×15×384 | | 2 | 48 | 192 | 192 | 111,024 |
| fire8 | 7×7×512 | 11×11×512 | 15×15×512 | | 2 | 64 | 256 | 256 | 188,992 |
| conv9 | 7×7×3 | 11×11×3 | 15×15×3 | 1×1/1 (×3) | 1 | | | | 1,539 |
| avgpool9 | 1×1×3 | 1×1×3 | 1×1×3 | 7×7/1 11×11/1 15×15/1 | 0 | | | | |
| Total #of parameters including batch normalization layers : 536,143 | | | | | | | Total # of parameters | | 526,531 |

## C. Fully convolutional networks and determination of optimal threshold

The existing methods [3], [4] convert the input images into non-overlap grids before extracting patches. This processing method will be referred to as the grid method for convenience. Furthermore, existing methods determine whether the fingerprint is fake or not by voting after determining the fake patches. This method is shown in Fig. 4 (a). In contrast, the proposed method processes like an FCN without extracting patches in the test, although it uses the same grid method in training [35]. The processing results appear in 3D tensors of fake, live, and background, as shown in Fig. 4 (b), and the liveness of the fingerprint is determined from the probabilities of being live or fake, excluding results classified as background.

The final determination about the liveness of a fingerprint is made by comparing the determined liveness probability with the optimal threshold obtained from the training data. The FCN method can process more evaluations of fake fingerprints than the grid method. Furthermore, all fingerprints can be processed without segmentation, because the background is included in the classes for the classification.

The optimal threshold was determined as follows. Among the 3D tensors obtained by the FCN model, those with the highest background probability are removed. The other live and fake probabilities are normalized. This is equal to the probability of liveness in the local position of fingerprints, excluding background. The average liveness probability at each position obtained in this way becomes the final liveness score of the fingerprint. This process is described in Algorithm 1.

The method in Fig. 4 (b) determines the optimal threshold as a value in the training data that has the minimum average classification error (ACE). After executing Algorithm 1 for all training data, the liveness probabilities are calculated using the

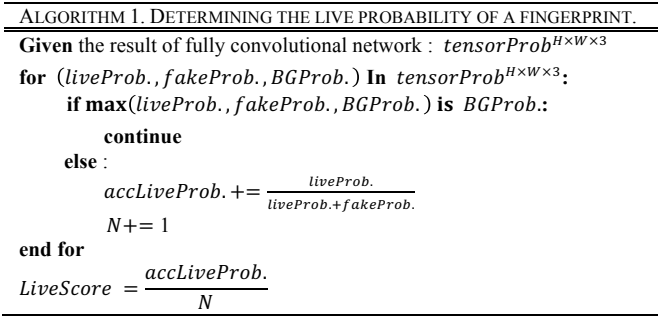

**ALGORITHM 1. DETERMINING THE LIVE PROBABILITY OF A FINGERPRINT.**

**Given** the result of fully convolutional network : $tensorProb^{H \times W \times 3}$
**for** $(liveProb., fakeProb., BGProb.)$ **In** $tensorProb^{H \times W \times 3}$:
    **if** $\max(liveProb., fakeProb., BGProb.)$ **is** $BGProb.$:
        **continue**
    **else** :
        $accLiveProb. += \frac{liveProb.}{liveProb.+fakeProb.}$
        $N += 1$
**end for**
$LiveScore = \frac{accLiveProb.}{N}$

FCN. If the value calculated through Algorithm 1 is $s$, the process of obtaining the optimal threshold is expressed by the following equations:

$$Ferrfake(t) = \int_t^1 P(s|fake)ds \tag{7}$$

$$Ferrlive(t) = \int_0^t P(s|live)ds \tag{8}$$

$$t^* = \arg\min_t \frac{Ferrfake(t) + Ferrlive(t)}{2} \tag{9}$$

Equation (7) means the probability where a fake fingerprint is classified as a live fingerprint, because *LiveScore s* calculated through Algorithm 1 is higher than threshold $t$. Equation (8) is the opposite; that is, the probability of classification of a live fingerprint as a fake fingerprint. Equation (9) finds $t^*$ at which the averages of *Ferrfake* and *Ferrlive* are the lowest. This $t^*$ value is applied when a fake fingerprint is tested.

## IV. EXPERIMENT RESULTS

### A. Dataset

The data used to evaluate the proposed method were LivDet2011 [1], LivDet2013 [6], and LivDet2015 [7]. With LivDet2013, only the data of the Biometrika sensor and Italdata sensors obtained by the non-cooperative method were used. A fingerprint image obtained from the Swipe sensor of LivDet2013 is different from the existing data because it is obtained by swiping the fingerprint from top to bottom. The Crossmatch sensor of LivDet2013 was excluded because it showed a problem when acquiring fingerprints [36]. The LivDet data used in this experiment are outlined in Table III.

The LivDet2015 data include data consisting of materials that were not used in training, so only the data that consist of fake fingerprint materials used in training were used in the test.

TABLE III
LIVDET DATA USED IN THE EXPERIMENT

| | Sensor | Size | DPI | # of testing (Live/Fake) | # of Fake materials |
|---|---|---|---|---|---|
| LivDet11 | Biometrika | 312×372 | 500 | 1000/1000 | 5 |
| | Digital | 355×391 | 500 | 1000/1000 | 5 |
| | Italdata | 640×480 | 500 | 1000/1000 | 5 |
| | Sagem | 352×384 | 500 | 1000/1000 | 5 |
| LivDet13 | Biometirka | 312×372 | 569 | 1000/1000 | 5 |
| | Italdata | 640×480 | 500 | 1000/1000 | 5 |
| LivDet15 | Biometirka | 1000×1000 | 1000 | 1000/1000 | 4 |
| | Crossmatch | 800×750 | 500 | 1500/851 | 3 |
| | Digital | 252×324 | 500 | 1000/1000 | 4 |
| | Greenbit | 500×500 | 500 | 1000/1000 | 4 |

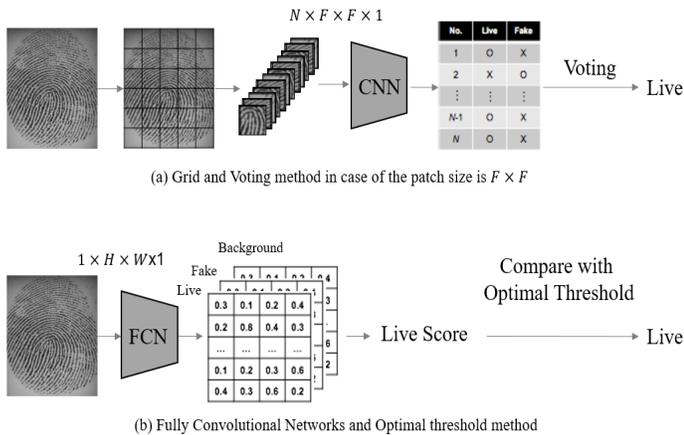

(a) Grid and Voting method in case of the patch size is $F \times F$

(b) Fully Convolutional Networks and Optimal threshold method

Fig 1. Comparison between the grid and voting method and fully convolutional and optimal threshold method.

As a result, the numbers of live and fake fingerprints are different in the case of the Crossmatch sensor of LivDet2015.

### B. Experimental environment

The NVIDIA GTX 1080 GPU was used for training and testing the CNN. For the parameters applied to training, a learning rate of 0.0005, a batch size of 64, and an epoch number of 60 were used. For the optimizer, Adamax [37] was applied, and 10% of the training LivDet data were used for validation. When extracting validation data from the training data, the fingerprints for training and validation were divided before extracting the patches from them to prevent the inclusion of patches obtained from the same fingerprint in both training and validation. For the training patches, random horizontal flip was applied in batch units. The loss of validation data was checked for each epoch in the learning process, and if the loss did not decrease over four epochs, the learning rate was reduced to half. The hyper-parameters used in this experiment are listed in Table IV.

### C. Comparison of performance between the grid method and the fully convolutional neural network method

The ACE used is as shown in (10) to compare the performance between the grid method and the FCN method. As explained above about the method of finding the optimal threshold, ACE is the average error detection rate of live fingerprints ($Ferrlive$) and the error detection rate of fake fingerprints ($Ferrfake$):

$$ACE = (Ferrlive + Ferrfake)/2 \qquad (10)$$

For comparison, in the grid method, patches with the highest background probability were removed from the results of each patch classification result and the same process of finding the optimal threshold was applied. Then this value was used to classify the fake fingerprints. The comparison between the grid and FCN methods is shown in Table V and Fig. 5. State of the art (SOTA) is a method using the patch-based CNN by Wang et al. [3] and the VGG model, with no patch applied by Nougueira et al [25]. Among the results of SOTA LivDet2015 represents the performance of the award-winning result by Nougueira et al [7]. When compared with the performance of SOTA, the proposed method showed the lowest average ACE. This comparison confirmed that the FCN method exhibited better performance on average than the grid method. This verifies that the FCN method is a better choice when distinguishing fake fingerprints using patches. The results of this calculation are outlined in Table VI. The best performance

### TABLE IV
#### HYPER-PARAMETERS USED IN THE LEARNING OF TRANSFORMED SQUEEZENET

| GPU | NVIDIA GTX 1080 |
|---|---|
| Learning rate | 0.0005 (making it half when validation loss does not decrease for 4 epochs) |
| Batch Size | 64 |
| Epoch | 60 |
| Optimizer | Adamax |
| Validation data | 10% of training data |

### TABLE V
#### COMPARISON OF PERFORMANCE BETWEEN THE GRID AND FCN METHODS BY PATCH SIZE (ACE).

| Dataset | | | Patch Size 32×32 | | Patch Size 48×48 | | Patch Size 64×64 | |
|---|---|---|---|---|---|---|---|---|
| | | SOTA | Grid | FCN | Grid | FCN | Grid | FCN |
| 11 | Bio | 3.5 [3] | 2.3 | 2.35 | 2.05 | 1.1 | 3.35 | 1.55 |
| | Dig | 0 [3] | 0.7 | 0.9 | 0.75 | 1.1 | 0.85 | 0.8 |
| | Ita | 0 [3] | 6 | 5.4 | 5.4 | 4.75 | 5.35 | 4.1 |
| | Sag | 0 [3] | 1.63 | 1.09 | 0.99 | 1.56 | 1.19 | 1.34 |
| 13 | Bio | 0.8 [25] | 0.15 | 0.15 | 0.3 | 0.35 | 0.2 | 0.2 |
| | Ita | 0 [3] | 0.55 | 0.4 | 0.5 | 0.4 | 0.65 | 0.65 |
| 15 | Bio | 5.6 [7] | 1.3 | 1.25 | 1 | 0.35 | 1.4 | 0.6 |
| | Cro | 1.53 [7] | 1 | 0.82 | 1.82 | 1.09 | 1.91 | 1.44 |
| | Dig | 6.35 [7] | 3.7 | 3 | 4.1 | 3.4 | 5.7 | 5.45 |
| | Gre | 3.9 [7] | 0.4 | 0.3 | 0.4 | 0.2 | 0.85 | 0.55 |
| **Average** | | **2.34** | **1.77** | **1.57** | **1.73** | **1.43** | **2.15** | **1.67** |

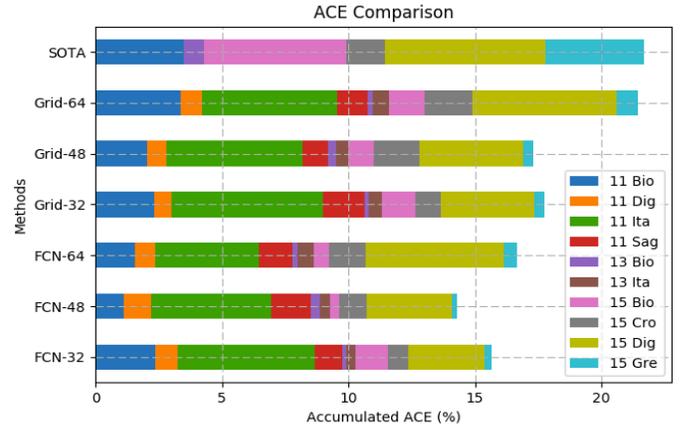

Fig 2. Comparison of ACE between Grid and FCN by patch size.

### TABLE VI
#### COMPARISON OF PERFORMANCE BETWEEN GRID AND FCN BY PATCH SIZE (ACE).

| Patch Size | **Grid** | **FCN** | **Avg. over Patch size** |
|---|---|---|---|
| 32×32 | 1.77 | 1.57 | **1.67** |
| 48×48 | 1.73 | 1.43 | **1.58** |
| 64×64 | 2.15 | 1.67 | **1.91** |
| **Avg. over Model** | **1.88** | **1.56** | |

was obtained on average when the FCN method was applied and the patch size was 48×48.

The processing speed between the grid and FCN methods was compared, and the results are shown in Table VII and Fig. 6. The grid method extracts patches without allowing any overlapping of fingerprints, and these patches are grouped into batches for calculation. The FCN method is calculated as a general convolution type, and the final results are represented in a three-dimensional tensor. When calculating processing speed, only the processing time of the model in the GPU was compared, while excluding the time required for extracting patches from the grid method and the time for calculating the final optimal threshold score.

| Patch Size | | 32×32 | | 48×48 | | 64×64 | |
|---|---|---|---|---|---|---|---|
| | Image Size | Grid | FCN | Grid | FCN | Grid | FCN |
| 11 Bio | 372×312 | 66 | 50 | 50 | 50 | 36 | 50 |
| Dig | 391×355 | 84 | 59 | 59 | 59 | 51 | 59 |
| Ita | 480×640 | 172 | 124 | 148 | 125 | 128 | 125 |
| Sag | 384×352 | 82 | 58 | 59 | 58 | 51 | 58 |
| 13 Bio | 372×312 | 67 | 50 | 49 | 50 | 36 | 50 |
| Ita | 480×640 | 171 | 124 | 146 | 125 | 128 | 125 |
| 15 Bio | 1000×1000 | 532 | 397 | 411 | 397 | 399 | 398 |
| Cro | 750×800 | 312 | 239 | 248 | 239 | 239 | 239 |
| Dig | 324×252 | 49 | 37 | 31 | 37 | 29 | 37 |
| Gre | 500×500 | 136 | 102 | 114 | 102 | 86 | 102 |
| **Average** | | **167** | **124** | **131** | **124** | **118** | **124** |

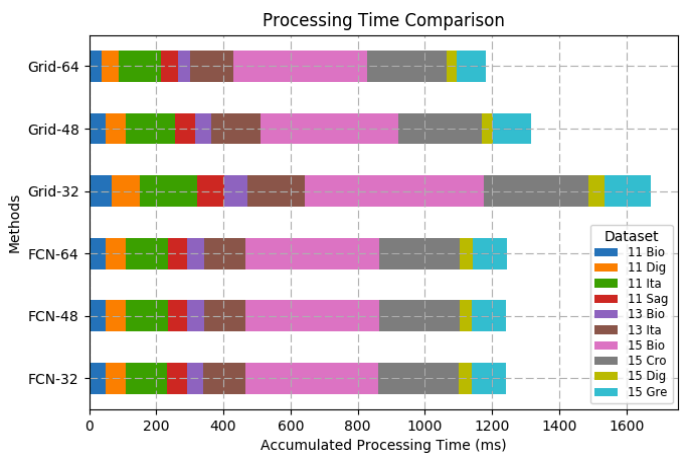

Fig 3. Comparison of processing speed between Grid and FCN.

The FCN applying the proposed optimal threshold can process a fingerprint in around 124 ms, on average. With the FCN method, there was no difference in processing speed by patch, because only the filter size of the average pooling layer, which is the last layer, is different by patch size. With the grid method, as the patch size increases, the number of patches acquired from fingerprints decreases, but the spatial size of patches increases. The grid method needs fewer calculations when the number of acquired patches decreases than when the spatial size of a patch increases. The FCN method is a better choice, because it is faster and has a better ACE, compared to the grid method, except for the 64×64 model.

### D. Comparison of performance between the voting method and the optimal threshold method

The proposed method determines the optimal threshold of patches from the training data. In this experiment, performance was compared between the results obtained by determining the optimal threshold and the voting method, which simply counts the classified patches. The comparison results are outlined in Table VIII and Fig. 7. Because the optimal threshold can be applied to the grid method as well, this result is also shown. The results shown in the "Thres." column in Table VIII are identical to the results in Table VII, which have been inserted for comparison. In Fig. 7, Vot-Grid-64 indicates the grid method of the 64×64 patch to which the voting method was applied, and Thres-FCN-64 indicates the FCN method of the 64×64 patch to which the optimal threshold was applied. The results suggest that for both the grid and FCN methods, applying the optimal threshold shows better detection performance.

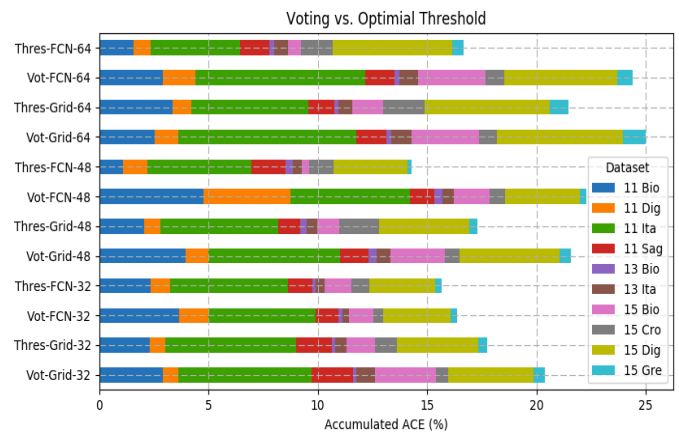

Fig 4. Comparison between the voting method and the optimal threshold method.

| | | Patch Size: 32×32 | | | | Patch Size: 48×48 | | | | Patch Size: 64×64 | | | |
| | | Grid | | **FCN** | | Grid | | **FCN** | | Grid | | **FCN** | |
| Data Set | | Voting | Thres. | Voting | Thres. | Voting | Thres. | Voting | Thres. | Voting | Thres. | Voting | Thres. |
| 11 | Bio | 2.9 | 2.3 | 3.65 | 2.35 | 3.95 | 2.05 | 4.75 | 1.1 | 2.55 | 3.35 | 2.9 | 1.55 |
| | Dig | 0.7 | 0.7 | 1.35 | 0.9 | 1.05 | 0.75 | 4 | 1.1 | 1.05 | 0.85 | 1.5 | 0.8 |
| | Ita | 6.1 | 6 | 4.9 | 5.4 | 6 | 5.4 | 5.45 | 4.75 | 8.15 | 5.35 | 7.75 | 4.1 |
| | Sag | 1.9 | 1.63 | 1.04 | 1.09 | 1.33 | 0.99 | 1.14 | 1.56 | 1.37 | 1.19 | 1.37 | 1.34 |
| 13 | Bio | 0.15 | 0.15 | 0.2 | 0.15 | 0.35 | 0.3 | 0.35 | 0.35 | 0.25 | 0.2 | 0.2 | 0.2 |
| | Ita | 0.85 | 0.55 | 0.3 | 0.4 | 0.65 | 0.5 | 0.55 | 0.4 | 0.9 | 0.65 | 0.85 | 0.65 |
| 15 | Bio | 2.8 | 1.3 | 1.1 | 1.25 | 2.5 | 1 | 1.6 | 0.35 | 3.1 | 1.4 | 3.1 | 0.6 |
| | Cro | 0.57 | 1 | 0.43 | 0.82 | 0.66 | 1.82 | 0.71 | 1.09 | 0.84 | 1.91 | 0.84 | 1.44 |
| | Dig | 3.9 | 3.7 | 3.1 | 3 | 4.55 | 4.1 | 3.45 | 3.4 | 5.75 | 5.7 | 5.2 | 5.45 |
| | Gre | 0.5 | 0.4 | 0.3 | 0.3 | 0.55 | 0.4 | 0.3 | 0.2 | 1.05 | 0.85 | 0.7 | 0.55 |
| **Average** | | **2.04** | **1.77** | **1.64** | **1.57** | **2.16** | **1.73** | **2.23** | **1.43** | **2.5** | **2.15** | **2.44** | **1.67** |

## E. Comparison with a model using data augmentation

The model of the FCN method applying the proposed optimal threshold was compared with the model that was trained by applying data augmentation, and the results are outlined in Table IX. Data augmentation is a method that applies a vertical flip in addition to the random horizontal flip applied in the basic model. Table IX outlines the FCN results quoted from tables V and VIII, and Aug-FCN indicates the model that applied data augmentation. These results show that data augmentation with a vertical flip does not improve performance, on average, and the effect of data augmentation varies by sensor. When only the average ACE is compared, the FCN method trained with 48×48 patches, which had shown the best performance in the past, improved performance through data augmentation, with an ACE of 1.35%, which was the best in the experiments so far. Fig. 8 is a graphic representation of the cumulative ACE of Table IX. Fig. 9 shows the detection error tradeoff (DET) curve of the model that was trained through data augmentation with the 48×48 patch, which showed the best performance among the proposed methods.

## F. Generalization performance of the proposed method

Experiments until now used fake fingerprints created with the same type of material for both training and testing. However,

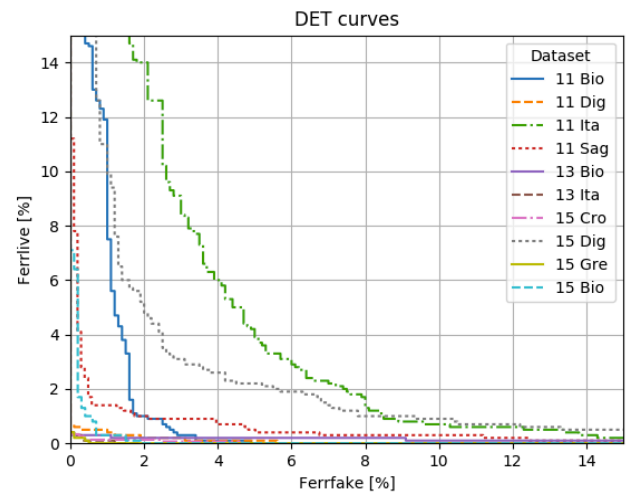

Fig. 5. DET curve of the proposed method.

attacks with fake fingerprints created with unknown materials that have not been used for training can happen. LivDet2015 contains fake fingerprints made of unknown materials in the test data. Fake fingerprint materials that do not exist in the training data are liquid ecoflex and RTV for Biometrika, Digital Persona, and Green Bit sensors, and OOMOO and gelatin for Crossmatch sensors.

Table X outlines the fake fingerprint–detection rates for unknown materials using the FCN method that applied the optimal threshold trained through data augmentation, which showed the highest ACE (1.35%) among the proposed methods. For comparison with other algorithms, the fake fingerprint–detection rate (Fcorrfake) of the same method, as presented in LivDet2015 [7], is also shown in Table X. A total of 12 teams and individuals participated in the competition, and their results are listed together for comparison with the proposed method. The comparison results confirm that the detection rate of the proposed method is not bad, even for data that was not used in training.

TABLE IX
COMPARISON OF PERFORMANCE WITH THE MODEL APPLYING DATA AUGMENTATION (ACE)

| Patch Size | | 32×32 | | 48×48 | | 64×64 | |
|---|---|---|---|---|---|---|---|
| Dataset | | FCN | Aug FCN | FCN | Aug FCN | FCN | Aug FCN |
| | Bio | 2.35 | **5.2** | 1.1 | **1.85** | 1.55 | **2.05** |
| 1 1 | Dig | 0.9 | **0.9** | 1.1 | **0.4** | 0.8 | **1** |
| | Ita | 5.4 | **4.3** | 4.75 | **4.65** | 4.1 | **5.95** |
| | Sag | 1.09 | **1.29** | 1.56 | **1.28** | 1.34 | **1.43** |
| 1 3 | Bio | 0.15 | **0.35** | 0.35 | **0.3** | 0.2 | **0.15** |
| | Ita | 0.4 | **1.4** | 0.4 | **0.6** | 0.65 | **0.45** |
| | Bio | 1.25 | **1.15** | 0.35 | **0.85** | 0.6 | **0.65** |
| 1 5 | Cro | 0.82 | **2.54** | 1.09 | **0.42** | 1.44 | **1.45** |
| | Dig | 3 | **2.9** | 3.4 | **2.95** | 5.45 | **5.75** |
| | Gre | 0.3 | **0.3** | 0.2 | **0.2** | 0.55 | **1.1** |
| Average | | **1.57** | **2.03** | **1.43** | **1.35** | **1.67** | **2** |

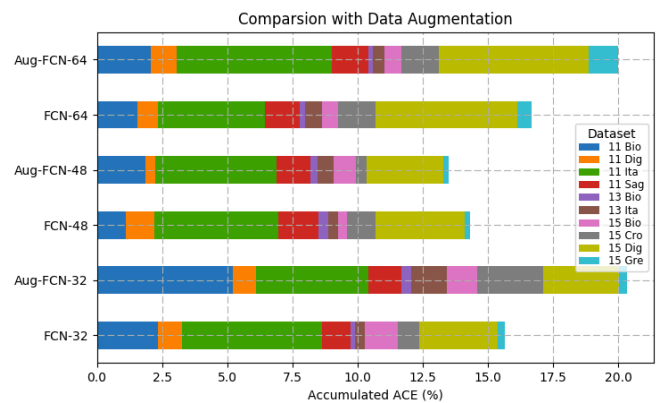

Fig. 6. Comparison with the model applying data augmentation.

TABLE X
FAKE FINGERPRINT–DETECTION RATE (%) FOR DATA NOT USED IN TRAINING

| | Biometrika | | Crossmatch | | Digital Persona | | Green Bit | |
|---|---|---|---|---|---|---|---|---|
| | Algorithm | Fcorr fake | Algorithms | Fcorr fake | Algorithms | Fcorr fake | Algorithms | Fcorr fake |
| 1 | unina | 98.6 | **Ours** | **100** | unina | 99.4 | unina | 96 |
| 2 | titanz | 95 | COPI LHA | 98.32 | nogu eira | 94 | nogu eira | 92.6 |
| 3 | nogu eira | 94.2 | anon ym | 95.98 | **Ours** | **89** | jingli an | 92.2 |
| 4 | hbirk holz | 93.8 | nogu eira | 95.98 | UFP E II | 85.4 | **Ours** | **92** |
| 5 | jingli an | 93.2 | jingli an | 88.44 | hbirk holz | 85.2 | titanz | 87.6 |
| 6 | CSI_ MM | 88.6 | unina | 86.1 | titanz | 84.4 | hecto rn | 87.2 |
| 7 | anon ym | 85.4 | hbirk holz | 81.41 | jingli an | 80.6 | anon ym | 86.4 |
| 8 | hecto rn | 83.2 | titanz | 80.74 | CSI | 75.8 | UFP E II | 83.6 |
| 9 | **Ours** | **83.2** | hecto rn | 76.55 | CSI_ MM | 73.2 | CSI_ MM | 82.2 |
| 10 | CSI | 80.8 | CSI_ MM | 70.18 | UFP E II | 72.4 | hbirk holz | 81.6 |
| 11 | UFP E II | 72 | CSI | 69.68 | hecto rn | 70.8 | CSI | 76 |
| 12 | UFP E I | 58.6 | UFP E I | 52.43 | anon ym | 70.8 | COPI LHA | 75.6 |
| 13 | COPI LHA | 57.2 | UFP E II | 45.9 | COPI LHA | 69.4 | UFP E I | 63 |

### G. Visualization of detection result

Fig. 10 shows the visualization of some fingerprints that generated errors in the model where the average ACE was the lowest at 1.35%. This figure shows the feature map output before the fake fingerprints were distinguished by applying the optimal threshold, which has been resized to the original image size through bilinear interpolation. Therefore, the visualization image appears more crushed than the original fingerprint image. The live, fake, and background probabilities are visualized in red, green, and blue channels, respectively. The first one in parentheses after the sensor name indicates the true label, and the second one indicates the classified result. This visualization is possible because the FCN method is applied, and the data are classified into the three classes of fake, live, and background, instead of only two classes of fake and live. Because the visualization of a misclassified fingerprint has a large portion of red and green parts, it is easy to see at which part the fake and live fingerprints are differentiated.

Fake fingerprints usually have black or white parts where the contours of fingerprints appear unnatural or crushed. Although the properties are slightly different by sensor type, there are cases where fake fingerprints can be distinguished with the naked eye. In the case of 2011 Sagem in Fig. 10, the black part tends to appear more greenish, but it was misclassified due to an overall high ratio of classification as a live fingerprint. With the 2015 Digital Persona, the part with a poor quality fingerprint is expressed in green, which indicates a fake. With 2013 Italdata, the fake of the original fingerprint can easily be seen with the naked eye because the outer boundary of the fingerprint that appears in fake fingerprints can clearly be seen in a large portion of the image. This part was distinguished as fake when it was visualized, but misdetection occurred due to a high overall ratio of classification as a live fingerprint. This is the limitation with methods that perform post-processing after training through patches. With the 2015 Green Bit, there was a residual fingerprint in the background, which was expressed as green, indicating a fake.

## V. CONCLUSIONS

Fingerprint biometrics can authenticate users with comparative ease, but it has the disadvantage of being easily faked. Fake fingerprint–detection methods can be classified as hardware-based or software-based. The hardware-based method is accurate, but has the disadvantage of high cost due to the additional hardware, whereas the software-based method is relatively inaccurate, but has the advantage of low cost. In this study, a CNN was applied to improve the inaccurate detection performance of the software-based method. However, a CNN is difficult to use in actual applications, because it generally requires a large number of parameters.

To solve this problem, this paper proposed a patch-based FCN method for fake fingerprint detection that applies an optimal threshold with a small number of parameters. The proposed method showed better results than existing methods in the experiments, such as the following.

- Fingerprints are classified into live, fake, and background, instead of only two classes of live and fake, and pre-processing, such as segmentation, is integrated into the CNN.
- The appropriate patch size for training was determined by experimenting with 32×32, 48×48, and 64×64 patches.
- The FCN was applied without cutting out the patches during testing, thus, decreasing the misdetection rate and increasing the processing speed.
- Instead of the voting method for final determination of fake fingerprints, an optimal threshold is applied, which decreased the misdetection rate.
- The CNN structure was designed with approximately 2.0 MB of network parameters, allowing it to work without the need for a large memory.
- The detection results can be analyzed through visualization of the results obtained through the FCN.

The proposed method automatically extracts patches in 32×32, 48×48, and 64×64 sizes during training, and learns through the CNN with a small number of parameters using the Fire module from SqueezeNet. CNN models have been made according to each patch size to enable measurement of performance by patch size.

The CNN models operate in the FCN method and find the optimal threshold from the training data. In the same way as the method for determining the optimal threshold during training, the trained FCN is applied to the fingerprint image, and a fake fingerprint is finally determined by comparing the calculated score with the optimal threshold.

The experiment results are outlined in Table XI. In every methodology, the optimal threshold method showed better performance, compared with the voting method. Furthermore, when the same optimal threshold method was applied, the FCN method showed better performance than the grid method. Both FCN and Aug-FCN in Table XI apply the optimal threshold method. Aug-FCN was trained through data augmentation. The proposed method showed the best performance when the patch size is 48×48. The FCN method also has a faster processing speed, and the proposed network has a very small number of parameters, which require around 2.0 MB of memory.

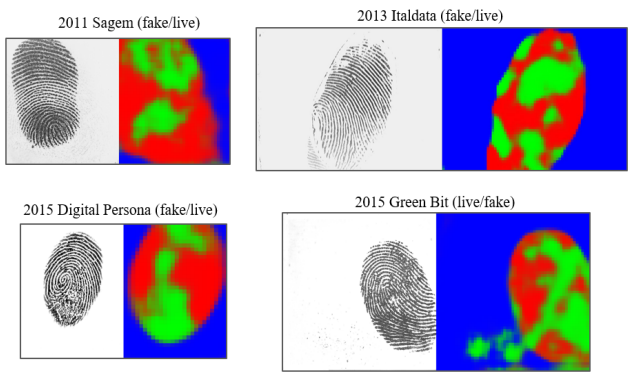

Fig 7. Visualization examples of misclassified fingerprints. Red, green, and blue indicate live, fake, and background, respectively.

TABLE XI
COMPARISON OF MODELS THAT SHOWED THE HIGHEST ACE

|  | Voting-Grid (32×32) | Thres-Grid (48×48) | FCN (48×48) | Aug-FCN (48×48) |
|---|---|---|---|---|
| ACE (%) | 2.04 | 1.73 | 1.43 | 1.35 |
| Processing Time (ms) | 167 | 131 | 124 | 124 |
| # of parameters | 536,143 (about 2.0MB) | | | |

The proposed method showed better methodologies in fake fingerprint detection than the existing CNN method using patches. In the future, to improve the LivDet data, which is slightly insufficient for training a CNN, such methods as a generative adversarial network will be applied, and the improvement in generalization performance will be researched to increase performance for unknown materials.